\documentclass{article}

\usepackage{microtype}
\usepackage{graphicx}
\usepackage{subfigure}
\usepackage{booktabs} 

\usepackage{hyperref}

\usepackage[accepted]{hill2019}

\usepackage{booktabs} 
\usepackage{ dsfont } 
\usepackage{amsmath}  
\usepackage{natbib} 
\usepackage{amsfonts} 

\DeclareMathOperator{\Pro}{Pr}

\newcommand{\logistic}[1]{f(#1)}
\newcommand{\sprod}[2]{\langle#1,#2\rangle}
\newcommand{\dd}[2]{\frac{d#1}{d#2}}

\icmltitlerunning{Aggregation of pairwise comparisons with reduction of biases}

\begin{document}
\twocolumn[
	
	\icmltitle{Aggregation of pairwise comparisons with reduction of biases}
	
	\begin{icmlauthorlist}
		\icmlauthor{Nadezhda Bugakova}{ya}
		\icmlauthor{Valentina Fedorova}{ya}
		\icmlauthor{Gleb Gusev}{ya}
		\icmlauthor{Alexey Drutsa}{ya}
	\end{icmlauthorlist}

	\icmlaffiliation{ya}{Yandex, Moscow, Russia}
	
	\icmlcorrespondingauthor{Nadezhda Bugakova}{belatriss@yandex-team.ru}
	\icmlcorrespondingauthor{Valentina Fedorova}{valya17@yandex-team.ru}

	\icmlkeywords{pairwise comparisons, aggregation, crowdsourcing, ranking}
	
	\vskip 0.3in
]
	
	\printAffiliationsAndNotice{\icmlEqualContribution}

	\begin{abstract}
	    We study the problem of ranking from crowdsourced pairwise comparisons.
	    Answers to pairwise tasks are known to be affected by the position of items on the screen~\cite{day:1969}, however, previous models for aggregation of pairwise comparisons do not focus on modeling such kind of biases.  
	    We introduce a new aggregation model factorBT for pairwise comparisons, which accounts for certain factors of pairwise tasks that are known to be irrelevant to the result of comparisons but may affect workers' answers due to perceptual reasons.
	    By modeling biases that influence workers, factorBT is able to reduce the effect of biased pairwise comparisons on the resulted ranking.
        Our empirical studies on real-world data sets showed that factorBT produces more accurate ranking from crowdsourced pairwise comparisons than previously established models.

	\end{abstract}

\section{Introduction}

Modern machine learning algorithms 
 require large amounts of labeled data to be trained.
Crowdsourcing marketplaces, such as Amazon Mechanical Turk\footnote{\url{https://mturk.com/}}, Figure Eight\footnote{\url{https://www.figure-eight.com}}, and Yandex Toloka\footnote{\url{https://toloka.yandex.com/}}, 
make it possible to obtain labels for large data sets in a shorter time as well as at a lower cost comparing to that needed for a limited number of experts. However, as workers at the marketplaces are non-professional and vary in levels of expertise, such labels are much noisier than those obtained from experts~\cite{snow/etal:2008}.
In order to reduce the noisiness, typically, each object is labelled by several workers, and then these labels are further aggregated in a certain way to infer a more reliable \emph{aggregated label} for the object~\cite{ipeirotis/etal:2014}.


Ranking problem implies recovering a total ordering over a set of objects, for example, documents or images. The problem is important for ranking-based applications like web search and recommender systems. To evaluate such algorithms preferences over objects are obtained from human workers. Since sets of objects needed to be ranked are large (e.g., for a given query search engines retrieve thousands of objects), workers are asked to express their preferences either in the form of \emph{absolute judgments} by grading each object on a given scale or in the form of \emph{pairwise comparisons} by choosing a winner among two objects. \citet{carterette/etal:2008} showed that in the latter approach workers perform tasks quicker and achieve a better inter-worker agreement.
This, in turn, motivates the problem of ranking from pairwise comparisons, which is addressed in this paper.


There are several sources of noise that can be observed in crowdsourced data in different tasks. A most studied kind of noise appears in multi--classification tasks, where workers, being non-professional, can confuse classes. A special case is when noisy labels come from spammers, which provide deterministic labels or answer randomly \cite{vuurens/etal:2011}. A large part of these workers can be automatic bots~\cite{difallah/etal:2012}, some others are real users trying to earn more money in short time~\cite{carterette/soboroff:2010}. In some cases, workers can be even malicious: they can deliberately provide incorrect labels for some reasons~\cite{raykar/yu:2012}.
To infer the true label in the presence of such type of noise, different unsupervised consensus models were proposed~\citep[e.g.,][]{dawid/skene:1979,whitehill/etal:2009,zhou/etal:2015}. 
These approaches are able to detect and implicitly eliminate or correct the noisy part of crowdsourced information.
Another kind of noise can be generated due to stochastic nature of true labels. In pairwise comparisons, a weaker player can beat a stronger one by chance. A document that is more preferable by the majority of users can be less preferable for some minority people, which are not neither spammers, nor malicious users. In this case, true labels are often modeled as random variables rather than deterministic values. For example, in the Bradley--Terry model~\cite{bradley/terry:1952}, the probability to win for a player is the logistic transformation of the difference between latent scores (skills) of the player and the opponent. The random outcome of a match is considered as the (unknown) true label, which can be spoiled in the crowdsourced setting by low-quality workers (see, e.g. the Crowd-BT model~\cite{chen/etal:2013}).

In this paper, we study the problem of ranking from pairwise comparisons in the presence of a novel type of noise, which was overlooked in previous studies on consensus modeling. We consider the {\it biases} of workers caused by some {\it known factors} that are irrelevant to the task and cannot a-priori influence the unknown true answer, but can actually bias noisy crowdsourced labels. Examples of such factors are the positions of the two objects on the screen of the worker, the background design of each document, when their relevance should be compared, and etc.
We propose factorBT, a new model for ranking from pairwise comparisons, which takes into account and eliminates individual biases of workers towards irrelevant factors. FactorBT includes Bradly-Terry model as a special case. 
To the best of our knowledge, this work is \emph{the first one addressing the problem of ranking aggregation from pairwise comparisons in the presence of multiple factors that are irrelevant to the task but can bias the results of pairwise comparisons}.


\section{Related work}\label{sec:rw}
Classical score-based probabilistic models~\citep{bradley/terry:1952,thurstone:1927} for ranking from pairwise comparisons were designed for comparisons without noise. 
Many previous studies for ranking from noisy pairwise comparisons~\citep{chen/etal:2013,volkovs/zemel:2012,sunahase/baba/kashima:2017,shah/wainwright:2017} were based on the assumption that noise occurred in crowdsourced labels is independent of pairwise tasks properties, in contrast in this paper we focus on modelling the dependence of noise on tasks factors that are a priori known to be irrelevant to the result of comparisons.

Several works modeled biases of workers towards certain classes for multi-classication~\citep{dawid/skene:1979,raykar/yu:2012,zhou/etal:2015,welinder/perona:2010} and ordinal labels~\citep{joachims/kartik:2016,lyu:2018,kao/etal:2018}. 
Some works modeled in-batch bias  \citep{zhuang/etal:2015, zhuang/young:2015} which occurs when each worker receives a batch of tasks and for these reason labels for tasks may be affected by other tasks in the same batch. 
In this work, we focus on modelling of another source of bias that is caused by some known features of pairwise task such as position bias discussed below.

It was observed long ago that results of pairwise comparisons are affected by position bias \cite{day:1969}. 
\citet{xu/etal:2016} focused on filtering out workers having position bias rather than on aggregating pairwise comparisons. For this reason, their approach to optimization process is not traditional, and we do not see an easy extension of this process to account for multiple factors of bias. Secondly, in their method, all labels from biased workers are discarded and therefore we waste some unbiased labels produced by workers, who are biased only when they are confused or tired. Both of these shortcomings are covered by our work.

There have been several works using factors of tasks for better aggregation of crowdsourced multiclass labels~\citep{ruvolo/whitehill/movellan:2013,jin/etal:2017,welinder/perona:2010,kajino/etal:2012, kamar/karpor/horwitz:2015, wauthier/jordan:2011}. 
In contrast to our work, all these works considered factors that are predictive of the true labels for tasks, whereas we consider factors that should not affect the result of comparisons. Secondly, all the models were designed for multiclass labels and can not obtain ranking from pairwise comparisons. For this reason models for aggregating multiclass labels have not been used in previous works on ranking from noisy pairwise comparisons \citep{chen/etal:2013,sunahase/baba/kashima:2017,xu/etal:2016}.

\section{Problem setup}\label{sec:problem}
Let  a set of \(N\) items \(D = \{d_1,...,d_N\}\) and a set of \(K\) workers \(W = \{w_1,...,w_K\}\) be given. 
We assume that each worker has completed some number of tasks to make a \emph{pairwise comparison} of two items from \(D\). 
Namely, in such a task, a worker \(w_k\in W\) receives two different items \(d_i\) and \(d_j\in D\), compares them on its own, and chooses one of the two answers \(i \succ_k j\) or \(j \succ_k i\) (where \( i \succ_k j\) represents that \(w_k\) prefers \(d_i\) over \(d_j\)). 
Formally, when all tasks are completed, we have the set of all pairwise comparisons made by workers in \(W\) denoted by \(P =\{(w_k, d_i, d_j):  i \succ_k j\}\). 
Note that, 
first, \(P\) may not include all possible pairs of items, 
second, the number of comparisons produced by different workers may vary,
and, third, the same pair of items can be compared by several workers. The latter is a standard approach to reduce the amount of noise that presents in the pairwise comparisons when they are obtained via crowdsourcing  where reliability of workers is unknown~\cite{chen/etal:2013}.

Assume that each task to compare documents \(d_i\) and \(d_j \in D\) given to a worker \(w_k \in W\) is described by a vector of \(M\) \emph{features} \(x_{w_k, d_i, d_j} \in \mathbb{R}^M\), for brevity further on it is written as \(x_{kij}\). Denote the collection of the features' vectors for all tasks in \(P\) as \(X_P = \{x_{kij}\}\). The features describe the presence of particular properties of the task that should not affect the result of the comparison when it is made by a perfect worker. 
However, in our crowdsourcing setting, non-professional workers may unconsciously be affected by such features of tasks (because of perceptual reasons) and may thus provide biased answers~\cite{day:1969, xu/etal:2016}. In other words, \emph{biased} comparisons demonstrate the reaction of workers to features of a given pairwise task rather than reflecting the true preference over items in the task.
For example, the features may include the position of compared items on the screen\footnote{e.g., which one of the two items is located to the left of the other one}, which is known as \emph{position bias}~\citep{day:1969, xu/etal:2016}.
To give examples of other features of tasks that may affect pairwise comparisons, consider the application of comparing the results of two search engines over a pool of queries. It is desirable 
that comparisons of two search engines results pages (SERPs) for queries assess the relevance of the SERPs content and should not be affected by the order of the results from the two search engines, the SERPs style, or the existence of non-relevant but colorful objects on them. 

In this paper, we consider the problem of constructing a ranking list of the items $D$ by means of \emph{score-based models} for pairwise comparisons \cite{bradley/terry:1952, thurstone:1927}. 
Such a model assumes that each item \(d_i\in D\) has a latent ``quality'' \emph{score} \(s_i \in  \mathbb{R}\) and models the probability of preferring an item over another one as a specific function of the difference between their scores. For example, in the Bradley-Terry (BT) model, the probability that an item \(d_i\in D\) will be preferred in a comparison over an item \(d_j \in D\) (denoted by $i\succ j$) is
\begin{equation}\label{eq:BT}
\Pro(i\succ j) = \logistic{s_i - s_j},
\end{equation}
where \(\logistic{x} := \frac{1}{1 + e^{-x}}\). The set of scores for all items in \(D\) is denoted as $S = \{s_i\}_{i = 1,\ldots,N}$. Traditionally, scores \(S\) are inferred by \emph{maximizing the log-likelihood} of the observed pairwise comparisons \(P\). For the BT model, the log-likelihood of the observed data is
\begin{equation}\label{eq:BT_ll}
\sum_{(w_k, d_i, d_j)\in P}\!\!\!\!\!\!\log \Pro(i \succ_k j) = \!\!\!\!\!\!\sum_{(w_k, d_i, d_j)\in P}\!\!\!\!\!\!\log \logistic{s_i - s_j}.
\end{equation}

Given the scores $S$ of the items $D$, one can sort the items according to their scores and obtain a \emph{ranking} \(\pi\) of items in \(D\), which is a permutation \(\pi: D \to \{1,\ldots,N\}\) mapping each item \(d_i\) to its rank \(\pi(i)\). 
We follow \cite{chen/etal:2013} and say that estimated scores of items are \emph{accurate} if they agree with \emph{ground truth quality} \(g_i \in \mathbb{R}\) of these items in the following sense. Given a set \(\{g_i\}\) of ground truth quality for items \(i \in D_g \subseteq D\), the  \emph{accuracy} is measured by 
\begin{equation}\label{eq:accuracy}
  {\rm ACC} = \frac{\sum\limits_{d_i,d_j\in D_g} \mathbb{I}(g_i > g_j \wedge s_i > s_j)}{\sum\limits_{d_i,d_j\in D_g}  \mathbb{I}(g_i > g_j)},  
\end{equation}
where the indicator function \(\mathbb{I}(A)\) equals to 1 when \(A\) is true and 0 otherwise.

Given the observed results of pairwise comparisons \(P\) and features \(X_P\) describing pairwise tasks presented to workers, the goal of this paper is to estimate the scores of items in \(D\) maximizing the likelihood of \(P\). In this paper, we address the following questions for the problem of estimating scores of items from noisy pairwise comparisons: Firstly, how to model pairwise comparisons with multiple types of bias using the features of pairwise tasks? Secondly, can we use the features of pairwise tasks for debiasing the results of pairwise comparisons and to obtain more accurate scores of items?

\section{Baselines}\label{sec:baselines}
In this section we briefly describe established score-based methods for ranking from noisy pairwise  comparisons and provide references to  papers describing the methods.
\subsection{Bradley-Terry model}
The classical Bradley-Terry (BT) model \cite{bradley/terry:1952}, described in Section~\ref{sec:problem}, is the basic approach for ranking from pairwise comparisons. For this model all workers are treated equally. To estimate unknown scores \(S\) for items the log-likelihood~\eqref{eq:BT_ll} of \(P\) is maximized.    
As expected, in the crowdsourcing setting where reliability of workers vary, the classical BT model performs badly~\cite{chen/etal:2013}.

\subsection{CrowdBT model}\label{subsec:crowdbt}
\citet{chen/etal:2013} described an extension  of the BT model to the crowdsourcing setting, where a parameter $\eta_k \in [0, 1]$ for each worker \(k \in W\) describes the worker's qualification. For their  CrowdBT model the probability that a worker $w_k$ prefers an item $d_i$ over another item $d_j$  is defined as follows: 
\[\Pro(i \succ_k j) = \eta_k \logistic{s_i-s_j} + (1 - \eta_k)\logistic{s_j-s_i}.\]

To optimize parameters in the case of sparse pairwise comparisons the authors suggested to use the virtual node regularization: 
the set of pairwise comparisons \(P\) is extended to $P_0$ by adding fictive comparisons of the virtual item $d_0$ with a score $s_0$ to all other items performed by a perfect worker \(w_0\) whose parameter $\eta_0 = 1$ and resulting in one virtual win for $d_0$ and one virtual loss: $P_0 = P \cup \{(w_{0}, d_0, d_i)_{i=1..N}\} \cup \{(w_{0}, d_i, d_0)_{i=1..N}\}$. 
The model parameters \(S\) and \(\eta = \{\eta_k\}_{k = 1,\ldots,K}\) are estimated by maximizing the sum
\begin{align*}
    \sum\limits_{k = 1}^K\sum\limits_{(w_k, d_i, d_j) \in P}\!\!\!\!\!\!\log\left(\eta_k \logistic{s_i-s_j} + (1 - \eta_k)\logistic{s_j-s_i}\right) + \\\lambda\sum\limits_{i = 1}^N \left(\log\logistic{s_0-s_i} + \log\logistic{s_i-s_0}\right), 
\end{align*}
where the first term is the  log-likelihood of \(P\), the second one is the regularization, and $\lambda$ is the regularization parameter.
CrowdBT by construction can not detect biased workers and consider them to be random: $\eta_k$ for such workers would be near 0.5 and log-likelihood addends corresponding to these workers become constant. Thus, CrowdBT cannot extract any information from biased answers. 

\textbf{4.3. Pairwise HITS}\\
\citet{sunahase/baba/kashima:2017} proposed a method called {\it pairwise HITS}, where
the relationships between scores \(S\) and ability \(r_k \in \mathds{R}\) of workers $w_k \in W$ are described by two equations. For any $d_j, d_{j'} \in D$:
\begin{equation}\label{eq:q_up}
s_j - s_{j'} = \sum_{k\in V_{j,j'}} r_k - \sum_{k\in V_{j',j}} r_k, 
\end{equation}
where \(V_{j,j'} = \{k = 1,\ldots, K \mid (w_k, d_j, d_{j'}) \in P\}\) is the set of comparisons in \(P\) for which \(j\) beats \(j'\).
These equations summarizing all comparisons in \(P\) can be rewritten in a matrix form and then items scores are estimated using a standard method. Workers' ability are approximated as the fraction of correct comparisons according to the current estimation of scores \(S\):
\begin{equation}\label{eq:r_up}
r_k = \frac{|\{(w_k, d_j, d_{j'}) \in P \mid s_j > s_{j'}\}|}{|\{(w_k, d_j, d_{j'}) \in P \}|}.
\end{equation}
Values for $\{r_k\}_{k=1,\ldots,K}$ and $S$ are updated according to equations~\eqref{eq:q_up},~\eqref{eq:r_up} 
alternately until convergence.

\textbf{4.4. Linear model}\\
A linear model incorporating position bias was proposed in~\cite{xu/etal:2016}. 
For each comparison $(w_k, d_i, d_j) \in P$ the result of this comparison $Y^k_{i, j}$ is equal to 1 if $d_i \succ_k d_j$ and -1 otherwise. The answer is modeled based on scores of items $S$ and noise: 
\[Y^k_{i, j} = s_i - s_j + \gamma^k + \epsilon_{ij}^k,\] 
 where $\epsilon_{ij}^k$ is random sub-gaussian noise and $\gamma^k$ is position bias specific for each worker \(k\). 
The problem can be rewritten in a matrix form and then LASSO approach is used for approximating optimal parameters.
Though the method tries to capture position bias, it is able to model only linear relations between parameters and our experiments in Sec.~\ref{sec:experiments} show that the model performs poorly in comparison with our method.

\section{FactorBT}\label{sec:factorBT}
In this section we describe our factorBT model for ranking from pairwise comparisons. 
For our model a worker $w_k \in W$ has parameters  $\gamma_k \in \mathds{R}$ and $r_k \in \mathds{R}^M$. The intuition for \(\gamma_k\) is the following: for a given pairwise task a worker \(w_k\) produces a non-biased answer (according to the BT model~\eqref{eq:BT}) with the probability $\logistic{\gamma_k}$ and with the probability $1 - \logistic{\gamma_k}$ her answer is affected by undesirable features of the task. The parameter $r_k \in \mathds{R}^M$ models the reaction of a worker \(w_k\) on tasks' features.
To model the strength of workers' reaction on tasks features we follow  \citet{ruvolo/whitehill/movellan:2013}. 
In our model, the amount of observed bias in answers of a worker depends on her sensitivity to certain tasks' features and on the presence of these features in tasks performed by this worker. 

In this way, for factorBT the probability that a worker $w_k \in W$ prefers an item \(d_i \in D\) over another item \(d_j \in D\) is
\begin{align*}
    \Pro(d_i \succ_k d_j) \!=\! \logistic{\gamma_k}\logistic{s_{i}\! -\! s_{j}} \!+\! (1 \!- \!\logistic{\gamma_k)}\logistic{\sprod{x_{kij}}{r_k}},
\end{align*}
where $\sprod{a}{b}$ is the scalar product of vectors \(a\) and \(b\).


 To estimate parameters parameters $\{\gamma_k\}_{k=1,\ldots,K}$, $\{r_k\}_{k=1,\ldots,K}$ and $S$ we maximize the log-likelihood of the observed pairwise comparisons \(P\). 
Similarly to other score based model, factorBT needs a regularization 
in the case when comparisons $P$ are sparse. For this purpose we use the virtual node regularization similarly to \citet{chen/etal:2013}, which has been described in Section~\ref{subsec:crowdbt}. This gives the following target function to maximize:

\vspace{-0.5cm}
\begin{small}
\begin{align*}
T = \sum\limits_{(w_k, d_i, d_j) \in P} \log[\logistic{\gamma_k}\logistic{s_{i} - s_{j}} +(1 - \logistic{\gamma_k})\times\\\times\logistic{\sprod{x_{kij}}{r_k}}] +\lambda\sum\limits_{i = 1}^N \log\logistic{s_i - s_0}+ \log \logistic{s_0 - s_i}.
\end{align*}
\vspace{-0.5cm}
\end{small}

To maximize the function we use a standard conjugate gradient descent algorithm. 
The gradients of $T$ for each parameter are the following:

\vspace{-0.5cm}
\begin{small}
\begin{align*}
\dd{T}{\gamma_k} =& \!\!\sum\limits_{(w_k, d_i, d_j) \in P}\!\! \frac{(\logistic{s_i - s_j} - \logistic{\sprod{r_k}{x_{kij}}})\logistic{\gamma_k}\logistic{-\gamma_k}}{\Pro(d_i \succ_k d_j)};\\
\dd{T}{r_k} =& \!\!\sum\limits_{(w_k, d_i, d_j) \in P}\!\!\frac{(1 - \logistic{\gamma_k})\logistic{\sprod{r_k}{x_{kij}}}\logistic{-\sprod{r_k}{x_{kij}}}x_{kij}}{\Pro(d_i \succ_k d_j)};\\
\dd{T}{s_i} =& \!\!\sum\limits_{(w_k, d_i, d_j) \in P}\!\! \frac{\logistic{\gamma_k} \logistic{s_i - s_j}\logistic{s_j - s_i}}{\Pro(d_i \succ_k d_j)} \\
 &\quad\quad\quad\quad +\lambda(\logistic{s_0 - s_i} - \logistic{s_i - s_0}), i = 1, .., N;\\
\dd{T}{s_0} =& \sum\limits_{d_i \in D} \lambda(\logistic{s_i - s_0} - \logistic{s_0 - s_i}).
\end{align*}
\end{small}
\vspace{-0.5cm}

Results of the optimization depend on the initial values for parameters. All scores $S^0$  (including $s_0$ for the virtual item) are initialized by 0. If a worker \(w_k\in W\) always answers the same answer her $\gamma_k$ is $f(-1)$ and $f(1)$ otherwise. The initialization of $r_k \in \mathds{R}^M$ is more complex. It can be explained using an example of our experimental data. In our experiments, tasks' features $x_{ijk}$ take values from $\{-1, 0, 1\}^M$. For each feature \(l \in {1, \dots M}\) (denoted as \(x_{kijl}\) for a given task \(w_k, d_i,d_j\)):  $1$  means that this feature presents in $d_i\in D$ (for example, a certain background design) and does not present in $d_j$;  $-1$ visa verse;  and $0$ means this feature either presents in both items or absents in both.
Then, let a vector $r_k$ consists of $M$ values of $r_{kl} \in \mathds{R}$. For each worker \(w_k\) and each vector component $r_{kl}$ we estimate smoothed statistics of her answers, i.e. the fraction of her answers for tasks with a feature \(l\) presenting: $a_{kl} = \frac{|\{ (w_k, d_i, d_j) \in P : x_{kijl} = 1\}|  + 1}{|\{(w_k, d_i, d_j) \in P: x_{kijl} = \pm1\}| + 2}$ . Finally, $r_{kl}$ is initialized by $\log(a_{kl})$\footnote{In factorBT \(r_k\) is used in \(\logistic{\sprod{r_k}{x_{kij}}} \), where \(x_{ijk}\in\{-1, 0, 1\}^M\), for this reason we apply logarithmic transformation for the initialisation.}

\section{Experiments}\label{sec:experiments}
\textbf{Simulated study.} In this experiment we follow the simulating procedure described in \citep[Section 5.1]{chen/etal:2013}. We assume that there are 100 objects and randomly generate unique 400 pairs to be used as pairwise tasks. Ground truth scores of items were randomly chosen without replacement among integers from 0 to 99. Every task (i.e. {\it the pair}) has two factors each of them is whether $1$, $0$ or $-1$ ($x \in \{-1, 0, 1\}^2$). 
The values of factors were chosen randomly with equal probability. The parameters $\gamma$ and $r_i, i = {1, 2}$ ($r \in \mathds{R}^2$) for 100 workers were sampled from the standard normal distribution $\mathcal{N}(0, 1)$. For each pair of items we randomly select 10 different workers to generate their pairwise comparisons and worker $j$ is voted for the first object with the probability $\Pro(d_1 \succ_j d_2) = \logistic{\gamma}\logistic{s_{d_1} - s_{d_2}} + (1 - \logistic{\gamma})\logistic{\sprod{x}{r}}$.

To evaluate the goodness of approximating true parameters by factorBT we use the following metrics: Firstly, as in \cite{chen/etal:2013} we compute accuracy of a ranking that was obtained by sorting objects based on estimated sores. The ranking generates golden true comparisons between each pair of elements and we can compute the accuracy~\eqref{eq:accuracy}. Secondly, as in \cite{whitehill/etal:2009} we compute paired Pearson correlation coefficient between true and approximated parameters of factorBT.
Results in Table~\ref{tab:sim} are averaged over 10 trials. It shows that correlation between reconstructed and original parameters is quite good and ranking itself is close to the ground truth one.
\vspace{-0.2in}
\begin{table}[ht]
	\caption{The goodness of estimating parameters for factorBT on a simulated data set.}
	\label{tab:sim}
\vskip -0.15in
\begin{center}
	\begin{small}
		\begin{sc}
\begin{tabular}{lr} 
	\toprule
Ranking accuracy & 0.51 \\ 
Pearson correlation for $r_1$ & 0.50 \\ 
Pearson correlation for $r_2$ & 0.47 \\ 
Pearson correlation for $\gamma$ & 0.81 \\ 
Pearson correlation for $s$ & 0.92 \\ 
\bottomrule
\end{tabular}
\end{sc}
\end{small}
\end{center}
\vskip -0.3in
\end{table}

\textbf{Reading difficulty data set. }
The data set~\footnote{We used the data set available at \url{http://www-personal.umich.edu/~kevynct/datasets/wsdm_rankagg_2013_readability_crowdflower_data.csv}.} contains 490 different documents (text passages). In each task a worker was given two passages of text (a passage A and a passage B). The task is to decide whether the passage A is more difficult to read and understand, than the passage B is more difficult or it was possible to answer that they can not decide. Pairs of passages for tasks were chosen randomly. Overall, 624 workers made 13856 comparisons (for 1424 tasks) but each worker contributes no more than 40 judgments. 
For some passages golden labels assessing their reading difficulty level on the scale from 1 to 12 were provided. 
Only determined answers (i.e. "Passage A is more difficult." and "Passage B is more difficult.") were used to compute accuracy of ranking~\eqref{eq:accuracy} for different methods.

\begin{figure}[ht]
    \vskip 0.2in
\begin{center}
\centerline{\includegraphics[width=\columnwidth]{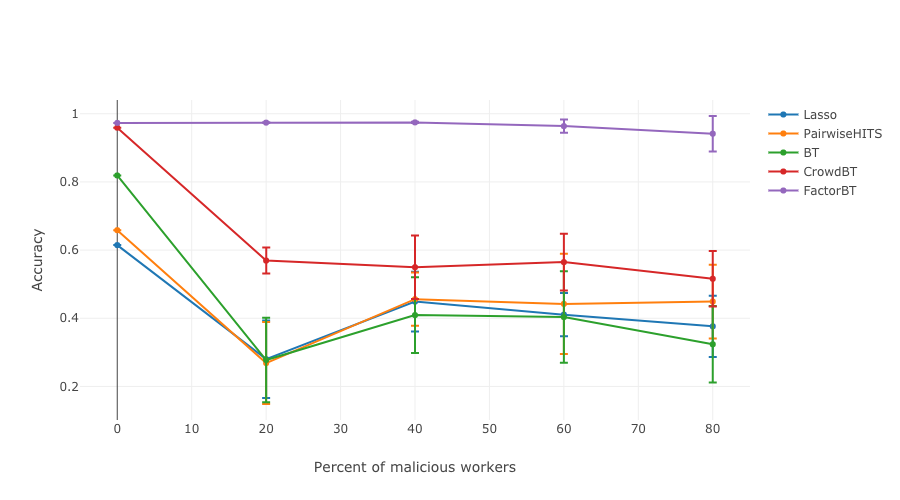}}
\caption{Accuracy for different methods for the reading difficulty data set. The horizontal axis shows percentage of added uniform spammers with left bias.}
\label{fig:wsdm_Left}
\end{center}
\vskip -0.2in
\end{figure}

Apart from the original data set we used its modifications where we add simulated malicious workers. It is known that workers solving pairwise tasks tend to choose left or first answer~\citep{day:1969, xu/etal:2016}, so for our simulation each malicious worker for each tasks always chooses a passage A as more difficult. This type of simulated workers is called uniform spammers according to~\cite{vuurens/etal:2011}. Tasks for each simulated worker are chosen randomly and the number of tasks per simulated worker is equal to the mean amount of tasks assigned to original workers (22 tasks for this data set). To evaluate stability of methods to uniform spammers, we compute accuracy of ranking obtained after adding different fractions of malicious workers from 0 to 100 percents (with the step 20) of the amount of original workers (i.e. \(0, 0.2K, \ldots, K \) of malicious workers are added if there are $K$ real ones). Accuracy results were averaged by 10 trials. The results are shown on Figure~\ref{fig:wsdm_Left}.

As we can see only FactorBT method is stable when uniform spammers appear in data  set. Moreover, even without spammers it shows the best result that means it detects position bias which is present in the data set.

\textbf{SERPs' comparisons data sets.}
Finally, we evaluated our algorithm on proprietary industry data. Each data set was collected to compare two systems using pairwise tasks. Two systems in each data set were chosen in a way that the underlying quality of one of them was a-priory known to be better than for another, and by the experimental design weaker systems had certain features that were irrelevant to the search quality but may attract workers attention. The task contains two SERP's of different search engines on the same query. Workers are asked to chose the SERP containing more relevant documents for a given query. Crowdsourced labels for the data sets were collected via Yandex.Toloka. 
Statistics about the data sets are summarized in Table~\ref{tab:info}. 
\vspace{-0.5cm}
\begin{table}[ht]
\caption{Statistics about proprietary data sets.}
\label{tab:info}
\vspace{-0.3cm}
\begin{center}
	\begin{small}
		\begin{sc}
\begin{tabular}{lcr} 
	\toprule
	& First & Second\\
	\midrule
Data set size & 2660 & 2500\\ 
Number of queries & 133 & 125\\ 
Number of workers & 194 & 194\\ 
Avg. number of tasks per worker & 13.7 & 20 \\ 
Avg. number of workers per task & 20 & 12.9\\
\bottomrule
\end{tabular}
\end{sc}
\end{small}
\end{center}
\vskip -0.1in
\end{table}
\vspace{-0.5cm}

The first data set compares results of {\it the first} page of Google and {\it the fifth} page of Yandex. The search engine's name is not shown to worker but certain workers are able to guess the search engine by the background design or other secondary features such as fonts, the distance between snippets, etc. As Yandex.Toloka is a product of Yandex apart of positional bias workers tend to have bias towards Yandex system. It is obvious though that the first page of Google should win.

The second data set contains Yandex first result page on a given query and Google first result page on a {\it spoiled query}. The idea is that we change the sense of query slightly so that results became less relevant than that for the original one. E.g., the original query is "Tom and Jerry" and the spoiled one is "Tom". 
In this case Yandex should win Google. 

Although for our proprietary data sets only SERPs are compared, we would like to obtain an \emph{overall score for the quality of each system}. For that purpose we compute the probability of one system to win another by averaging the probabilities based on estimated scores \(S\) of the first system's page win the second system's page over all queries in a data set. This is our main quality metric used for  experiments in this subsection.
To evaluate stability of methods to uniform spammers we evaluate quality scores for each system after adding different fractions of malicious workers, as described the previous experiment.

\begin{figure}[ht]
\begin{center}
\centerline{\includegraphics[width=\columnwidth]{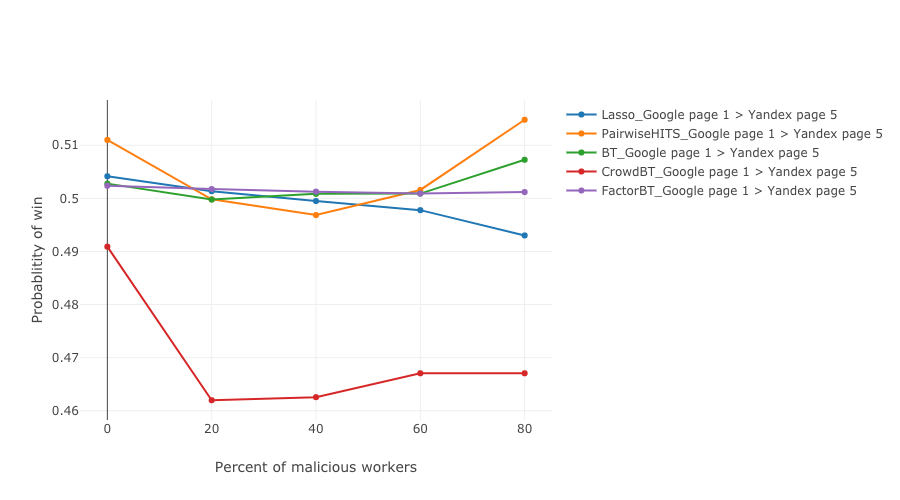}}
    \caption{\label{fig:part_2_Left} The overall quality score of the better system in the first data set. The horizontal axis shows percentage of added spammers with left bias.}
    \end{center}
\vskip -0.2in
\end{figure}

\begin{figure}[ht]
\begin{center}
\centerline{\includegraphics[width=\columnwidth]{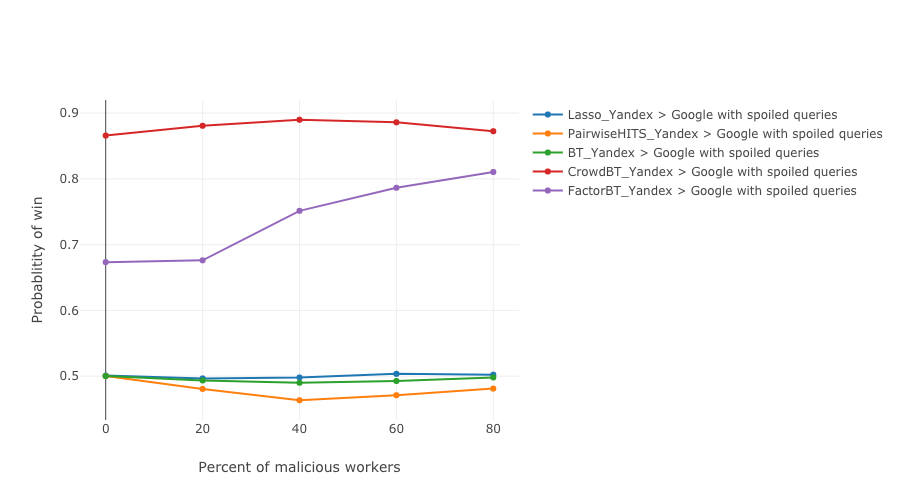}}
    \caption{\label{fig:part_4_Left_Yandex} The overall quality score of the better system in the second data set. The horizontal axis shows percentage of added spammers with both left and Yandex bias.}
    \end{center}
 \vskip -0.2in
\end{figure}

In the next figures we show probability of an a-priory better system to win an a-priory worse system, so if the line is above 0.5 then the overall quality of the better system was estimated correctly.
Figure~\ref{fig:part_2_Left} shows results for the first data set when we add uniform left-biased spammers. As we can see all methods but FactorBT are unstable when left spammers are added. Only FactorBT and BT show the probability more than 0.5 for all modifications of the data set. 
Figure~
\ref{fig:part_4_Left_Yandex} shows results for the second data set when we add mixture of uniform left-biased spammers and uniform Yandex-biased spammers (half of the spammers are of the first type and half of the second). Only FactorBT and CrowdBT show correct results on this data set. We noticed that CrowdBT looks more stable than FactorBT when the two kinds of spammers are added, though the former method does not model any biases. Also, we checked that if just one kind of spammers is added we observe similar stable results for FactorBT as for the first data set. So the situation with the two kind of spammers requires further study.
\vspace{-0.3cm}
\vspace{-0.7cm}
\vspace{-0.7cm}

\section{Conclusion}\label{sec:conclusion}
We have described the factorBT model for ranking from pairwise comparisons, which has the following properties: Firstly, it is a score-based model which, given a collection of pairwise comparisons, allows to obtain a ranking of items based on estimated scores for the items. Secondly, by modelling workers reaction to known irrelevant features presented in pairwise tasks it reduces the influence of these features on estimated scores of items.
Empirical evaluation with three real data sets has shown that: firstly, FactorBT produces more accurate ranking comparing to previously established baselines, and , secondly, factorBT shows a much stable performance to the addition of spammers.

\bibliographystyle{icml2019}
\bibliography{bibl}

\end{document}